\documentstyle[aclap,psfig,epsfig,times]{article}

\begin{document}

% for centering titles in tables
\newcommand{\mcc}[1]{\multicolumn{1}{c|}{#1}}

\newcommand{\example}[1]{\vspace{.7em} \\ \noindent \hspace*{0.0em}
\parbox{8cm}{\small #1}\vspace{0.9em}\\}

\newcommand{\interruptionpoint}{
\parbox[t]{0.7em}{
\raisebox{-1.8em}{\hspace{-.4em}\Huge\bf$\uparrow$} \vspace{-0.4em} \\
\makebox[1em][c]{\em interruption point} 
}}

\newcommand{\ip}[1]{
\parbox[t]{0.7em}{
\raisebox{-1.4em}{\LARGE\bf$\uparrow$} \vspace{-0.2em} \\
\makebox[0.3em][c]{\bf #1}}}

\newcommand{\reparandum}[1]{$\underbrace{\makebox{#1}}_{\makebox{\small\em reparandum}}$}
\newcommand{\editingterm}[1]{$\underbrace{\makebox{#1}}_{\makebox{\small\em
editing term}}$}
\newcommand{\alteration}[1]{$\underbrace{\makebox{#1}}_{\makebox{\small\em alteration}}$}

% used in eqnarray
\setlength{\arraycolsep}{1.5pt}

% format i-1 as a subscript
\newcommand{\im}{{\mbox{\tiny $i$-$1$}}}
\newcommand{\imt}{{\mbox{\tiny $i$-$3$}}}
\newcommand{\rim}{{\mbox{\tiny $1,\!i$-$1$}}}

\newcommand{\ignore}[1]{}
\newcommand{\pnote}[1]{\footnote{{\bf Note:} #1}}
\renewcommand{\pnote}[1]{}
\newcommand{\dash}{\mbox{\rm -}}

\title{INCORPORATING POS TAGGING INTO LANGUAGE MODELING\thanks 
{In proceedings of Eurospeech'97. This research work was 
completed while the first author was at the
University of Rochester. The authors would like to thank Geraldine
Damnati, Kyung-ho Loken-Kim, Tsuyoshi Morimoto, Eric Ringger and
Ramesh Sarukkai. This material is based upon work supported by the NSF
under grant IRI-9623665 and by ONR under grant N00014-95-1-1088.}}

\author{
Peter A. Heeman \\
France T\'el\'ecom CNET \\
Technopole Anticipa - 2 Avenue Pierre Marzin \\
 22301 Lannion Cedex, France. \\
{\tt heeman@lannion.cnet.fr}
\And 
James F. Allen \\
Department of Computer Science \\
University of Rochester \\
Rochester NY 14627, USA \\
{\tt james@cs.rochester.edu}}

\maketitle

\begin{abstract}
Language models for speech recognition tend to concentrate solely on
recognizing the words that were spoken.  In this paper, we redefine
the speech recognition problem so that its goal is to find both the
best sequence of words {\em and\/} their syntactic role (part-of-speech)
in the utterance.  This is a necessary first step towards tightening
the interaction between speech recognition and natural language
understanding.
\end{abstract}

%=====================
\section{INTRODUCTION}
%=====================

For recognizing spontaneous speech, the acoustic signal is to weak to
narrow down the number of word candidates.  Hence, speech recognizers
employ a language model that prunes out acoustic alternatives by
taking into account the previous words that were recognized.  In doing
this, the speech recognition problem is viewed as finding the most
likely word sequence $\hat{W}$ given the acoustic signal
\cite {Jelinek85}.
{\small\begin{eqnarray*}
\hat{W} & = & \arg\max_{W} \Pr(W|A) \\
        & = & \arg\max_{W} \frac{\Pr(A|W)\Pr(W)}{\Pr(A)} \\
        & = & \arg\max_{W} \Pr(A|W)\Pr(W)
\end{eqnarray*}}%
The last line involves two probabilities that need to be
estimated---the first due to the acoustic model $\Pr(A|W)$ and the
second due to the language model $\Pr(W)$.  The probability due to the
language model can be expressed as the following, where we rewrite the
sequence $W$ explicitly as the sequence of $N$ words $W_{1,N}$.
{\small\[ \Pr(W_{1,N}) = \prod_{i=1}^N \Pr(W_i|W_\rim) \]}%
To estimate the probability distribution, a training corpus is
typically used from which the probabilities can be estimated by
relative frequencies.  Due to sparseness of data, one must define
equivalence classes amongst the contexts $W_\rim$, which can be
done by limiting the context to an $n$-gram language model
\cite{Jelinek85} and also by grouping words into words classes
\cite{Brown-etal92:cl}.

Several attempts have been made to incorporate shallow syntactic
information to give better equivalence classes, where the shallow
syntactic information is expressed as part-of-speech (POS) tags
(e.g.~\cite{Jelinek85}, \cite{NieslerWoodland96:icassp}). A POS tag
indicates the syntactic role that a particular word is playing in the
utterance, e.g.~whether it is a noun or a verb, etc. The approach is
to use the POS tags of the prior few words to define the equivalence
classes. This is done by summing over all POS possibilities as shown
below.
{\small\begin{eqnarray*}
\lefteqn{\Pr(W_i|W_{1,\im})} \\
& = & \sum_{P_{1,i}} \Pr(W_i|P_{1,i}W_{1,\im})\Pr(P_{1,i}|W_{1,\im}) \\
& = & \sum_{P_{1,i}} \Pr(\!W_i|P_{1,i}W_{1,\im}\!)\Pr(\!P_i|P_{1,\im}W_{1,\im}\!)\Pr(\!P_{1,\im}|W_{1,\im}\!) \\
\end{eqnarray*}}%
Furthermore, the following two assumptions are made to simplify the
context.  
{\small\begin{eqnarray*}
\Pr(W_i|P_{1,i}W_{1,\im}) &\approx & \Pr(W_i|P_i) \\
\Pr(P_i|P_{1,\im}W_{1,\im})  &\approx & \Pr(P_i|P_{1,\im})
\end{eqnarray*}}%
However, this approach does not lead to an improvement in the
performance of the speech recognizer. For instance, Srinivas \cite
{Srinivas96:icslp} reports that such a model results in a 24.5\%
increase in perplexity over a word-based model on the Wall Street
Journal, and Niesler and Woodland \cite {NieslerWoodland96:icassp}
report an 11.3\% increase (but a 22-fold decrease in the number of
parameters of such a model). Only by interpolating in a word-based
model is an improvement seen \cite {Jelinek85}.

A more major problem with the above approach is that in a spoken
dialogue system, speech recognition is only the first step in
understanding a speaker's contribution.  One also needs to determine
the syntactic structure of the words involved, its semantic meaning,
and the speaker's intention in making the utterance.  This information
is needed to help the speech recognizer constrain the alternative
hypotheses.  Hence, we need a tighter coupling between speech
recognition and the rest of the interpretation process.

%===============================
\section{REDEFINING THE PROBLEM}
%===============================

As a starting point, we re-examine the approach of using POS tags in
the speech recognition process.  Rather than view POS tags as
intermediate objects solely to find the best word assignment, we
redefine the goal of the speech recognition process so that it finds
the best word sequence {\em and\/} the best POS interpretation given
the acoustic signal.
{\small\begin{eqnarray*}
\hat{W}\hat{P} & = & \arg\max_{WP} \Pr(WP|A) \\
               & = & \arg\max_{WP} \Pr(A|WP)\Pr(WP)
\end{eqnarray*}}%
The first term $\Pr(A|WP)$ is the acoustic model, which traditionally
excludes the category assignment. The second term $\Pr(WP)$ is the
POS-based language model. Just as before, we rewrite the probability of
$\Pr(WP)$ as a product of probabilities of the word and POS tag given
the previous context.
{\small\begin{eqnarray*}
\lefteqn{\Pr(W_{1,N}P_{1,N})} \\
 &=& \prod_{i=1,j} \Pr(W_iP_i|W_{1,\im}P_{1,\im}) \\
 &=& \prod_{i=1,j} \Pr(W_i|W_{1,\im}P_{1,i})\Pr(P_i|W_{1,\im}P_{1,\im}) 
\end{eqnarray*}}%
The final probability distributions are similar to those used for POS
tagging of written text \cite
{Charniak-etal93:aaai,Church88:anlp,DeRose88:cl}. However, these
approaches simplify the probability distributions as is done by
previous attempts to use POS tags in speech recognition language
models.\footnote{A notable exception is the work of Black {\em et
al.}~\cite{Black-etal92:darpa:pos}, who use a decision tree to learn
the probability distributions for POS tagging.} As we will show in
Section~\ref{sec:r:richness}, such simplifications lead to poorer
language models.

%=====================================
\section{ESTIMATING THE PROBABILITIES}
%=====================================

The probability distributions that we now need to estimate are more
complicated then the traditional ones. Our approach is to use the
decision tree learning algorithm \cite
{Bahl-etal89:tassp,Black-etal92:darpa:pos,Breiman-etal84:book}, which
uses information theoretic measures to construct equivalence classes of
the context in order to cope with sparseness of data. The decision
tree algorithm starts with all of the training data in a single leaf
node. For each leaf node, it looks for the question to ask of the
context such that splitting the node into two leaf nodes results in
the biggest decrease in {\em impurity}, where the impurity measures
how well each leaf predicts the events in the node. Heldout data is
used to decide when to stop growing the tree: a split is rejected if
the split does not result in a decrease in impurity with respect to
the heldout data. After the tree is grown, the heldout dataset is used
to smooth the probabilities of each node with its parent
\cite{Bahl-etal89:tassp}.

%---------------------------------------------
\subsection{Word and POS Classification Trees}
%---------------------------------------------

\begin{figure}
\begin{picture}(0,0)%
\includegraphics{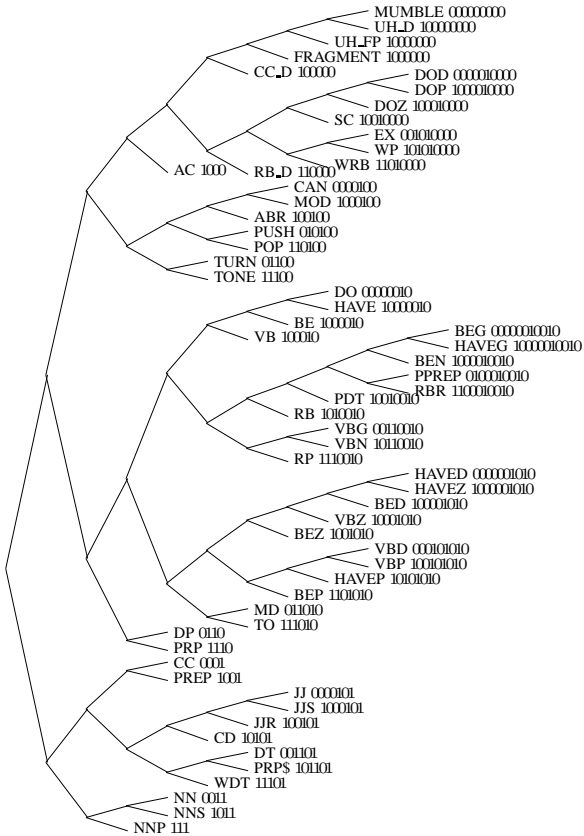}%
\end{picture}%
\setlength{\unitlength}{0.008750in}%
\begingroup\makeatletter\ifx\SetFigFont\undefined
% extract first six characters in \fmtname
\def\x#1#2#3#4#5#6#7\relax{\def\x{#1#2#3#4#5#6}}%
\expandafter\x\fmtname xxxxxx\relax \def\y{splain}%
\ifx\x\y   % LaTeX or SliTeX?
\gdef\SetFigFont#1#2#3{%
  \ifnum #1<17\tiny\else \ifnum #1<20\small\else
  \ifnum #1<24\normalsize\else \ifnum #1<29\large\else
  \ifnum #1<34\Large\else \ifnum #1<41\LARGE\else
     \huge\fi\fi\fi\fi\fi\fi
  \csname #3\endcsname}%
\else
\gdef\SetFigFont#1#2#3{\begingroup
  \count@#1\relax \ifnum 25<\count@\count@25\fi
  \def\x{\endgroup\@setsize\SetFigFont{#2pt}}%
  \expandafter\x
    \csname \romannumeral\the\count@ pt\expandafter\endcsname
    \csname @\romannumeral\the\count@ pt\endcsname
  \csname #3\endcsname}%
\fi
\fi\endgroup
\begin{picture}(265,505)(25,333)
\put(242,822){\makebox(0,0)[lb]{\smash{\SetFigFont{6}{7.2}{rm} MUMBLE $0\!0\!0\!0\!0\!0\!0\!0\!0\!$}}}
\put(242,813){\makebox(0,0)[lb]{\smash{\SetFigFont{6}{7.2}{rm} UH\_D $1\!0\!0\!0\!0\!0\!0\!0\!0\!$}}}
\put(218,804){\makebox(0,0)[lb]{\smash{\SetFigFont{6}{7.2}{rm} UH\_FP $1\!0\!0\!0\!0\!0\!0\!0\!$}}}
\put(194,795){\makebox(0,0)[lb]{\smash{\SetFigFont{6}{7.2}{rm} FRAGMENT $1\!0\!0\!0\!0\!0\!0\!$}}}
\put(170,786){\makebox(0,0)[lb]{\smash{\SetFigFont{6}{7.2}{rm} CC\_D $1\!0\!0\!0\!0\!0\!$}}}
\put(266,784){\makebox(0,0)[lb]{\smash{\SetFigFont{6}{7.2}{rm} DOD $0\!0\!0\!0\!0\!1\!0\!0\!0\!0\!$}}}
\put(266,775){\makebox(0,0)[lb]{\smash{\SetFigFont{6}{7.2}{rm} DOP $1\!0\!0\!0\!0\!1\!0\!0\!0\!0\!$}}}
\put(242,766){\makebox(0,0)[lb]{\smash{\SetFigFont{6}{7.2}{rm} DOZ $1\!0\!0\!0\!1\!0\!0\!0\!0\!$}}}
\put(218,757){\makebox(0,0)[lb]{\smash{\SetFigFont{6}{7.2}{rm} SC $1\!0\!0\!1\!0\!0\!0\!0\!$}}}
\put(242,748){\makebox(0,0)[lb]{\smash{\SetFigFont{6}{7.2}{rm} EX $0\!0\!1\!0\!1\!0\!0\!0\!0\!$}}}
\put(242,739){\makebox(0,0)[lb]{\smash{\SetFigFont{6}{7.2}{rm} WP $1\!0\!1\!0\!1\!0\!0\!0\!0\!$}}}
\put(218,730){\makebox(0,0)[lb]{\smash{\SetFigFont{6}{7.2}{rm} WRB $1\!1\!0\!1\!0\!0\!0\!0\!$}}}
\put(170,726){\makebox(0,0)[lb]{\smash{\SetFigFont{6}{7.2}{rm} RB\_D $1\!1\!0\!0\!0\!0\!$}}}
\put(122,727){\makebox(0,0)[lb]{\smash{\SetFigFont{6}{7.2}{rm} AC $1\!0\!0\!0\!$}}}
\put(194,717){\makebox(0,0)[lb]{\smash{\SetFigFont{6}{7.2}{rm} CAN $0\!0\!0\!0\!1\!0\!0\!$}}}
\put(194,708){\makebox(0,0)[lb]{\smash{\SetFigFont{6}{7.2}{rm} MOD $1\!0\!0\!0\!1\!0\!0\!$}}}
\put(170,699){\makebox(0,0)[lb]{\smash{\SetFigFont{6}{7.2}{rm} ABR $1\!0\!0\!1\!0\!0\!$}}}
\put(170,690){\makebox(0,0)[lb]{\smash{\SetFigFont{6}{7.2}{rm} PUSH $0\!1\!0\!1\!0\!0\!$}}}
\put(170,681){\makebox(0,0)[lb]{\smash{\SetFigFont{6}{7.2}{rm} POP $1\!1\!0\!1\!0\!0\!$}}}
\put(146,672){\makebox(0,0)[lb]{\smash{\SetFigFont{6}{7.2}{rm} TURN $0\!1\!1\!0\!0\!$}}}
\put(146,663){\makebox(0,0)[lb]{\smash{\SetFigFont{6}{7.2}{rm} TONE $1\!1\!1\!0\!0\!$}}}
\put(218,654){\makebox(0,0)[lb]{\smash{\SetFigFont{6}{7.2}{rm} DO $0\!0\!0\!0\!0\!0\!1\!0\!$}}}
\put(218,645){\makebox(0,0)[lb]{\smash{\SetFigFont{6}{7.2}{rm} HAVE $1\!0\!0\!0\!0\!0\!1\!0\!$}}}
\put(194,636){\makebox(0,0)[lb]{\smash{\SetFigFont{6}{7.2}{rm} BE $1\!0\!0\!0\!0\!1\!0\!$}}}
\put(170,627){\makebox(0,0)[lb]{\smash{\SetFigFont{6}{7.2}{rm} VB $1\!0\!0\!0\!1\!0\!$}}}
\put(290,631){\makebox(0,0)[lb]{\smash{\SetFigFont{6}{7.2}{rm} BEG $0\!0\!0\!0\!0\!0\!1\!0\!0\!1\!0\!$}}}
\put(290,622){\makebox(0,0)[lb]{\smash{\SetFigFont{6}{7.2}{rm} HAVEG $1\!0\!0\!0\!0\!0\!1\!0\!0\!1\!0\!$}}}
\put(266,613){\makebox(0,0)[lb]{\smash{\SetFigFont{6}{7.2}{rm} BEN $1\!0\!0\!0\!0\!1\!0\!0\!1\!0\!$}}}
\put(266,604){\makebox(0,0)[lb]{\smash{\SetFigFont{6}{7.2}{rm} PPREP $0\!1\!0\!0\!0\!1\!0\!0\!1\!0\!$}}}
\put(266,595){\makebox(0,0)[lb]{\smash{\SetFigFont{6}{7.2}{rm} RBR $1\!1\!0\!0\!0\!1\!0\!0\!1\!0\!$}}}
\put(218,590){\makebox(0,0)[lb]{\smash{\SetFigFont{6}{7.2}{rm} PDT $1\!0\!0\!1\!0\!0\!1\!0\!$}}}
\put(194,581){\makebox(0,0)[lb]{\smash{\SetFigFont{6}{7.2}{rm} RB $1\!0\!1\!0\!0\!1\!0\!$}}}
\put(218,572){\makebox(0,0)[lb]{\smash{\SetFigFont{6}{7.2}{rm} VBG $0\!0\!1\!1\!0\!0\!1\!0\!$}}}
\put(218,563){\makebox(0,0)[lb]{\smash{\SetFigFont{6}{7.2}{rm} VBN $1\!0\!1\!1\!0\!0\!1\!0\!$}}}
\put(194,554){\makebox(0,0)[lb]{\smash{\SetFigFont{6}{7.2}{rm} RP $1\!1\!1\!0\!0\!1\!0\!$}}}
\put(266,545){\makebox(0,0)[lb]{\smash{\SetFigFont{6}{7.2}{rm} HAVED $0\!0\!0\!0\!0\!0\!1\!0\!1\!0\!$}}}
\put(266,536){\makebox(0,0)[lb]{\smash{\SetFigFont{6}{7.2}{rm} HAVEZ $1\!0\!0\!0\!0\!0\!1\!0\!1\!0\!$}}}
\put(242,527){\makebox(0,0)[lb]{\smash{\SetFigFont{6}{7.2}{rm} BED $1\!0\!0\!0\!0\!1\!0\!1\!0\!$}}}
\put(218,518){\makebox(0,0)[lb]{\smash{\SetFigFont{6}{7.2}{rm} VBZ $1\!0\!0\!0\!1\!0\!1\!0\!$}}}
\put(194,509){\makebox(0,0)[lb]{\smash{\SetFigFont{6}{7.2}{rm} BEZ $1\!0\!0\!1\!0\!1\!0\!$}}}
\put(242,500){\makebox(0,0)[lb]{\smash{\SetFigFont{6}{7.2}{rm} VBD $0\!0\!0\!1\!0\!1\!0\!1\!0\!$}}}
\put(242,491){\makebox(0,0)[lb]{\smash{\SetFigFont{6}{7.2}{rm} VBP $1\!0\!0\!1\!0\!1\!0\!1\!0\!$}}}
\put(218,482){\makebox(0,0)[lb]{\smash{\SetFigFont{6}{7.2}{rm} HAVEP $1\!0\!1\!0\!1\!0\!1\!0\!$}}}
\put(194,473){\makebox(0,0)[lb]{\smash{\SetFigFont{6}{7.2}{rm} BEP $1\!1\!0\!1\!0\!1\!0\!$}}}
\put(170,464){\makebox(0,0)[lb]{\smash{\SetFigFont{6}{7.2}{rm} MD $0\!1\!1\!0\!1\!0\!$}}}
\put(170,455){\makebox(0,0)[lb]{\smash{\SetFigFont{6}{7.2}{rm} TO $1\!1\!1\!0\!1\!0\!$}}}
\put(122,450){\makebox(0,0)[lb]{\smash{\SetFigFont{6}{7.2}{rm} DP $0\!1\!1\!0\!$}}}
\put(122,441){\makebox(0,0)[lb]{\smash{\SetFigFont{6}{7.2}{rm} PRP $1\!1\!1\!0\!$}}}
\put(122,432){\makebox(0,0)[lb]{\smash{\SetFigFont{6}{7.2}{rm} CC $0\!0\!0\!1\!$}}}
\put(122,423){\makebox(0,0)[lb]{\smash{\SetFigFont{6}{7.2}{rm} PREP $1\!0\!0\!1\!$}}}
\put(194,414){\makebox(0,0)[lb]{\smash{\SetFigFont{6}{7.2}{rm} JJ $0\!0\!0\!0\!1\!0\!1\!$}}}
\put(194,405){\makebox(0,0)[lb]{\smash{\SetFigFont{6}{7.2}{rm} JJS $1\!0\!0\!0\!1\!0\!1\!$}}}
\put(170,396){\makebox(0,0)[lb]{\smash{\SetFigFont{6}{7.2}{rm} JJR $1\!0\!0\!1\!0\!1\!$}}}
\put(146,387){\makebox(0,0)[lb]{\smash{\SetFigFont{6}{7.2}{rm} CD $1\!0\!1\!0\!1\!$}}}
\put(170,378){\makebox(0,0)[lb]{\smash{\SetFigFont{6}{7.2}{rm} DT $0\!0\!1\!1\!0\!1\!$}}}
\put(170,369){\makebox(0,0)[lb]{\smash{\SetFigFont{6}{7.2}{rm} PRP\$ $1\!0\!1\!1\!0\!1\!$}}}
\put(146,360){\makebox(0,0)[lb]{\smash{\SetFigFont{6}{7.2}{rm} WDT $1\!1\!1\!0\!1\!$}}}
\put(122,351){\makebox(0,0)[lb]{\smash{\SetFigFont{6}{7.2}{rm} NN $0\!0\!1\!1\!$}}}
\put(122,342){\makebox(0,0)[lb]{\smash{\SetFigFont{6}{7.2}{rm} NNS $1\!0\!1\!1\!$}}}
\put( 98,333){\makebox(0,0)[lb]{\smash{\SetFigFont{6}{7.2}{rm} NNP $1\!1\!1\!$}}}
\end{picture}
\vspace{-0.5em}
\caption{POS Classification Tree}
\label{fig:postree}
\end{figure}

To allow the decision tree to ask about the words and POS tags in the
context, we cluster the words and POS tags using the algorithm of
Brown {\em et al.}~\cite{Brown-etal92:cl} into a binary classification
tree. The algorithm starts with each word (or POS tag) in a separate
class, and successively merges classes that result in the smallest
lost in mutual information in terms of the co-occurrences of these
classes. By keeping track of the order that classes were merged, we
can construct a hierarchical classification of the words.
Figure~\ref{fig:postree} shows a classification tree that we grew for
the POS tags.
%
% analword -i VAR=P,SMUSH,LEAFNAME,INDENT=24 -f Pos2Tree.lsp postree.input > postree.tree
% analline -f Tree2Fig.lsp -i COMPACT=0.7 postree.tree > postree.fig     
% go into fig, export as latex/postscript shrunk to 65%
%
The binary classification tree gives an implicit binary encoding for
each word and POS tag, which we show after each POS tag in the figure.
The decision tree algorithm can then ask questions about the binary
encoding of the words, such as `is the third bit of the POS tag
encoding equal to one?', and hence can ask about which partition a word
is in.

Unlike other work that uses classification trees as the basis for the
questions used by a decision tree (e.g.~\cite
{Black-etal92:darpa:pos}), we treat the word identities as a further
refinement of the POS tags. This approach has the advantage of
avoiding unnecessary data fragmentation, since the POS tags and word
identities will not be viewed as separate sources of information. We
grow the classification tree by starting with a unique class for each
word and each POS tag that it takes on. When we merge classes to form
the hierarchy, we only allow merges if all of the words in both
classes have the same POS tag. The result is a word classification
tree for each POS tag. This approach to growing the word trees
simplifies the task, since we can take advantage of the hand-coded
linguistic knowledge (as represented by the POS tags). Furthermore, we
can better deal with words that can take on multiple senses, such as
the word ``loads'', which can be a plural noun ({\bf NNS}) or a
present tense third-person verb ({\bf PRP}).\footnote{Words-POS
combinations that occur only once in the training corpus are grouped
together in the class {\bf $<$unknown$>$}, which is unique for each
POS tag.}

\begin{figure}
\hspace{1cm}
\begin{picture}(0,0)%
\includegraphics{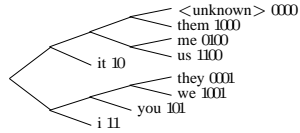}%
\end{picture}%
\setlength{\unitlength}{0.008750in}%
\begingroup\makeatletter\ifx\SetFigFont\undefined
% extract first six characters in \fmtname
\def\x#1#2#3#4#5#6#7\relax{\def\x{#1#2#3#4#5#6}}%
\expandafter\x\fmtname xxxxxx\relax \def\y{splain}%
\ifx\x\y   % LaTeX or SliTeX?
\gdef\SetFigFont#1#2#3{%
  \ifnum #1<17\tiny\else \ifnum #1<20\small\else
  \ifnum #1<24\normalsize\else \ifnum #1<29\large\else
  \ifnum #1<34\Large\else \ifnum #1<41\LARGE\else
     \huge\fi\fi\fi\fi\fi\fi
  \csname #3\endcsname}%
\else
\gdef\SetFigFont#1#2#3{\begingroup
  \count@#1\relax \ifnum 25<\count@\count@25\fi
  \def\x{\endgroup\@setsize\SetFigFont{#2pt}}%
  \expandafter\x
    \csname \romannumeral\the\count@ pt\expandafter\endcsname
    \csname @\romannumeral\the\count@ pt\endcsname
  \csname #3\endcsname}%
\fi
\fi\endgroup
\begin{picture}(97,84)(25,754)
\put(122,822){\makebox(0,0)[lb]{\smash{\SetFigFont{6}{7.2}{rm} $<$unknown$>$ $0\!0\!0\!0\!$}}}
\put(122,813){\makebox(0,0)[lb]{\smash{\SetFigFont{6}{7.2}{rm} them $1\!0\!0\!0\!$}}}
\put(122,804){\makebox(0,0)[lb]{\smash{\SetFigFont{6}{7.2}{rm} me $0\!1\!0\!0\!$}}}
\put(122,795){\makebox(0,0)[lb]{\smash{\SetFigFont{6}{7.2}{rm} us $1\!1\!0\!0\!$}}}
\put( 74,790){\makebox(0,0)[lb]{\smash{\SetFigFont{6}{7.2}{rm} it $1\!0\!$}}}
\put(122,781){\makebox(0,0)[lb]{\smash{\SetFigFont{6}{7.2}{rm} they $0\!0\!0\!1\!$}}}
\put(122,772){\makebox(0,0)[lb]{\smash{\SetFigFont{6}{7.2}{rm} we $1\!0\!0\!1\!$}}}
\put( 98,763){\makebox(0,0)[lb]{\smash{\SetFigFont{6}{7.2}{rm} you $1\!0\!1\!$}}}
\put( 74,754){\makebox(0,0)[lb]{\smash{\SetFigFont{6}{7.2}{rm} i $1\!1\!$}}}
\end{picture}
\vspace{-0.5em}
\caption{A Word Classification Tree}
\label{fig:prptree}
\end{figure}

In Figure~\ref{fig:prptree}, we give the classification tree for the
personal pronouns ({\bf PRP}). It is interesting to note that the
clustering algorithm distinguished between the subjective pronouns
`I', `we', and `they', and the objective pronouns `me', `us', and
`them'. The pronouns `you' and `it' can take either case, and the
algorithm partitioned them according to their most common usage in the
training corpus.
Although distinct POS tags could have been added to distinguish
between these two cases, it seems that the clustering algorithm can
make up for some of the shortcomings of the tagset.\footnote{The words
included in the {\bf $<$unknown$>$} class are the reflexive pronouns
`themselves', and `itself', which each occurred once in the training corpus.}

%-------------------------------
\subsection{Composite Questions}
%-------------------------------

In the previous section, we discussed the elementary questions that
can be asked of the words and POS tags in the context. However, there
might be a relevant partitioning of the data that can not be expressed
in that form. For instance, a good partitioning of a node might
involve asking whether questions $q_1$ and $q_2$ are both true. Using
elementary questions, the decision tree would need to first ask
question $q_1$ and then ask $q_2$ in the true subnode
created by $q_1$. This means that the false case has been split into
two separate nodes, which could cause unnecessary data fragmentation.

Unnecessary data fragmentation can be avoided by allowing composite
questions.  Bahl~{\it et al.}~\cite{Bahl-etal89:tassp} introduced a simple but
effective approach for constructing composite questions.  Rather than
allowing any boolean combination of elementary questions, they
restrict the typology of the combinations to {\em pylons}, which have
the following form ({\em true} maps all data into the true subset).
\begin{eqnarray*}
\mbox{\em pylon} & \Rightarrow & \mbox{\em true} \\
\mbox{\em pylon} & \Rightarrow & (\mbox{\em pylon} \wedge \mbox{\em elementary}) \\
\mbox{\em pylon} & \Rightarrow & (\mbox{\em pylon} \vee \mbox{\em elementary})
\end{eqnarray*}
The effect of any binary question is to divide the data into true and
false subsets. The advantage of pylons is that each successive
elementary question has the effect of swapping data from the true
subnode into the false or vice versa. Hence, one can compute the
change in node impurity that results from each successive elementary
question that is added. This allows one to use a greedy algorithm to
build the pylon by successively choosing the elementary question that
results in the largest decrease in node impurity.

We actually employ a beam search and explore the best 10 alternatives
at each level of the pylon. Again we make use of the heldout data to
help pick the best pylon, but we must be careful not to make too much
use of it for otherwise it will become as biased as the training data.
If the last question added to a candidate pylon results in an increase
in node impurity with respect to the heldout data, we remove that
question and stop growing that alternative. When there are no further
candidates that can be grown, we choose the winning pylon as the one
with the best decrease in node impurity with respect to the training
data. The effect of using composite questions is explored in
Section~\ref{sec:r:composites}.

%================
\section{RESULTS}
%================

To demonstrate our model, we have tested it on the Trains corpus
\cite{HeemanAllen95:cdrom}, a collection of human-human task-oriented
spoken dialogues consisting of 6 and half hours worth of speech, 34
different speakers, 58,000 words of transcribed speech, with a
vocabulary size of 860 words. To make the best use of the limited
amount of data, we use a 6-fold cross validation procedure, in which
we use each sixth of the corpus for testing data, and the rest for
training data.  

A way to measure a language model is to compute the {\em perplexity}
it assigns to a test corpus, which is an estimate of how well the
language model is able to predict the next word. The perplexity of a
test set of $N$ words $w_{1,N}$ is calculated as follows,
\[ 2^{-\frac1N\sum_{i=1}^N\log_2\hat{\Pr}(w_i|w_{1,i-1})} \]
where $\hat{\Pr}$ is the probability distribution supplied by the
language model. Full details of how we compute the word-based
perplexity are given in \cite{Heeman97:thesis}. We also measure the
error rate in assigning the POS tags. Here, as in measuring the
perplexity, we run the language model on the hand-transcribed word
annotations.

%------------------------------------
\subsection{Effect of Richer Context}
%------------------------------------
\label{sec:r:richness}

Table~\ref{tab:context} gives the perplexity and POS tagging error
rate (expressed as a percent).  To show the
effect of the richer modeling of the context, we vary the amount of
context given to the decision tree.  As shown by the perplexity
results, the context used for traditional POS-based language models
(second column) is very impoverished.  As we remove the simplifications to 
the context, we see the perplexity and POS tagging rates improve.  By using
both the previous words and previous POS tags as the context, we achieve
a 43\% reduction in perplexity and a 5.4\% reduction in the POS error rate.
\begin{table}[htb]
\setlength{\tabcolsep}{0.15em}
\small
\begin{center}
\begin{tabular}{|l|r|r|r|r|} \hline
Context for $W_i$ 	&\mcc{$P_{\mbox{\tiny $i$}}$}                 
			&\mcc{$P_{\mbox{\tiny $i$-3,$i$}}$}
			&\mcc{$P_{\mbox{\tiny $i$-3,$i$}}W_{\mbox{\tiny $i$-3,$i$-1}}$}
			&\mcc{$P_{\mbox{\tiny $i$-3,$i$}}W_{\mbox{\tiny $i$-3,$i$-1}}$} \\
Content for $P_i$	&\mcc{$P_{\mbox{\tiny $i$-3,$i$-1}}$}
			&\mcc{$P_{\mbox{\tiny $i$-3,$i$-1}}$}
                  	&\mcc{$P_{\mbox{\tiny $i$-3,$i$-1}}$}
			&\mcc{$P_{\mbox{\tiny $i$-3,$i$-1}}W_{\mbox{\tiny $i$-3,$i$-1}}$} \\ \hline
POS Error Rate            &  3.13 &  3.10 &  3.03 &  2.97 \\ 
Perplexity                & 42.32 & 32.11 & 29.49 & 24.17 \\ \hline
\end{tabular} \vspace{-0.8em}
\end{center}
\caption{Using Richer Contexts}
\label{tab:context}
\end{table}%

%------------------------------------------
\subsection{Constraining the Decision Tree}
%------------------------------------------

As we mentioned earlier, the word identity information $W_{i-j}$ is
viewed as further refining the POS tag of the word $P_{i-j}$.  Hence,
questions about the word encoding are only allowed if the POS tag is
uniquely defined.  Furthermore, for both POS and word questions, we
restrict the algorithm so that it only asks about more specific bits
of the POS tag and word encodings only if it has already uniquely
identified the less specific bits.  In Table~\ref{tab:constraints}, 
we contrast the effectiveness of adding further constraints.  The second
column gives the results of adding no further constraints, the third column
only allows questions about a POS tag $P_{i-j-1}$ only if $P_{i-j}$
is uniquely determined, and the fourth column adds the constraint
that the word $W_{i-j}$ must also be uniquely identified before questions
are allowed of $P_{i-j-1}$.
\begin{table}\begin{center}\begin{tabular}{|l|r|r|r|} \hline
                          &\mcc{None}&\mcc{POS}&\mcc{Full}\\ \hline
POS Error Rate            &     3.19 &    2.97 &   3.00  \\ 
Perplexity                &    25.64 &   24.17 &   24.39 \\ \hline
\end{tabular} \vspace{-0.8em}
\end{center}
\caption{Adding Additional Constraints}
\label{tab:constraints}
\end{table}%

From the table, we see that it is worthwhile to force the decision
tree to fully explore a POS tag for a word in the context before
asking about previous words. Hence, we see that the decision tree
algorithm needs help in learning that it is better to fully explore
the POS tags. However, we see that adding the further constraint that
the word identity should also be fully explored results in a decrease
in performance of the model. Hence, we see that it is not worthwhile
for the decision tree to fully explore the word information (which is
the basis of class-based approaches to language modeling), and it is
able to learn this on its own.

%--------------------------------
\subsection{Effect of Composites}
%--------------------------------
\label{sec:r:composites}

The next area we explore is the benefit of composite questions in
estimating the probability distributions. The second column of
Table~\ref {tab:composites} gives the results if composite questions
are not employed, the third column gives the results if composite
questions are employed, and the fourth gives the results if we employ
a beam search in finding the best pylon (with up to 10 alternatives).
From the results, we see that the use of pylons reduces the word
perplexity rate by 4.7\%, and the POS error rate by 2.3\%.
Furthermore, we see that using a beam search, rather than an entirely
greedy algorithm accounts for some of the improvement.
\begin{table}[h]
\begin{center} \begin{tabular}{|l|r|r|r|r|r|} \hline
                     	  & Not Used& Single&\mcc{10} \\ \hline 
POS Error Rate            &  3.04 &  3.04 &  2.97   \\ 
Perplexity                & 25.36 & 24.36 & 24.17   \\ \hline
\end{tabular} \vspace{-0.8em}
\end{center}
\caption{Effect of Composite Questions}
\label{tab:composites}
\end{table}%

%------------------------------------
\subsection{Effect of Larger Context}
%------------------------------------

In Table~\ref{tab:ngrams}, we look at the effect of the size of the
context, and compare the results to a word-based backoff language
model \cite{Katz87:assp} built using the CMU toolkit \cite
{Rosenfeld95:arpa}. For a bigram model, it has a perplexity of 29.3,
in comparison to our word perplexity of 27.4. For a trigram model, the
word-based model has a perplexity of 26.1, in comparison to our
perplexity of 24.2. Hence we see that our POS-based model results in a
7.2\% improvement in perplexity.
\begin{table}[h]
\setlength{\tabcolsep}{0.5em}
\begin{center}\small
\begin{tabular}{|l|r|r|r|} \hline
                          & Bigram&Trigram& 4-gram\\ \hline
POS Error Rate            &  3.19 &  2.97 &  2.97 \\ 
Perplexity                & 27.37 & 24.26 & 24.17 \\ \hline
Word-based Model          & 29.30 & 26.13 &       \\ \hline
\end{tabular} \vspace{-0.8em}
\end{center}
\caption{Using Larger Contexts}
\label{tab:ngrams}
\end{table}%

%===================
\section{CONCLUSION}
%===================

In this paper, we presented a new way of incorporating POS information
into a language model. Rather than treating POS tags as intermediate
objects solely for recognizing the words, we redefine the speech
recognition problem so that its goal is to find the best word sequence
{\em and} their best POS assignment. This approach allows us to use
the POS tags as part of the context for estimating the probability
distributions. In fact, we view the word identities in the context
as a refinement of the POS tags rather than viewing the POS tags and word
identities as two separate sources of information. To deal with this
rich context, we make use of decision trees, which can use information
theoretic measures to automatically determine how to partition the
contexts into equivalence classes. We find that this model results in
a 7.2\% reduction in perplexity over a trigram word-based model for
the Trains corpus of spontaneous speech. Currently, we are exploring
the effect of this model in reducing the word error rate.

Incorporating shallow syntactic information into the speech
recognition process is just the first step. In other work
\cite{Heeman97:thesis,HeemanAllen97:acl}, this syntactic information,
as well as the techniques introduced in this paper, are used to help
model the occurrence of dysfluencies and intonational phrasing in a
speech recognition language model. Our use of decision trees to
estimate the probability distributions proves effective in dealing
with the richer context provided by modeling these spontaneous speech
events. Modeling these events improves the perplexity to 22.5, a 14\%
improvement over the word-based trigram backoff model, and reduces the
POS error rate by 9\%.

\bibliographystyle{fullnamei}

\end{document}